\def\year{2021}\relax
\documentclass[letterpaper]{article} 
\usepackage{aaai21}  
\usepackage{times}  
\usepackage{helvet} 
\usepackage{courier}  
\usepackage[hyphens]{url}  
\usepackage{graphicx} 
\usepackage{natbib}  
\usepackage{caption} 
\usepackage{epsfig}
\usepackage{amsmath}
\usepackage{amssymb}
\usepackage{comment}
\usepackage{algorithm,algorithmicx,algpseudocode}
\usepackage[english]{babel}
\usepackage{subfigure}
\usepackage{algcompatible}
\usepackage{booktabs}
\usepackage{multirow}
\usepackage{eqparbox,array}

\usepackage[switch]{lineno}  %

\title{Ada-Segment: Automated Multi-loss Adaptation for Panoptic Segmentation}
\author{Gengwei Zhang\textsuperscript{\rm 1}, 
Yiming Gao\textsuperscript{\rm 1}, 
Hang Xu\textsuperscript{\rm 2}, 
Hao Zhang\textsuperscript{\rm 3}, 
Zhenguo Li\textsuperscript{\rm 2}, 
Xiaodan Liang\textsuperscript{\rm 1}\thanks{Corresponding Author: xdliang328@gmail.com} \\ 
}
\affiliations{
\textsuperscript{\rm 1}Sun Yat-Sen University 
\textsuperscript{\rm 2}Huawei Noah's Ark Lab 
\textsuperscript{\rm 3}Shanghai Jiao Tong University
}

\begin{document}
\maketitle

\begin{abstract}
    Panoptic segmentation that unifies instance segmentation 
    and semantic segmentation has recently attracted increasing attention. 
    While most existing methods focus on 
    designing novel architectures,
    we steer toward a different perspective: 
    performing automated multi-loss adaptation (named Ada-Segment) on the fly to flexibly adjust multiple 
    training losses over the course of training using a controller trained to capture the learning dynamics. 
    This offers a few advantages: 
    it bypasses manual tuning of the sensitive loss combination,
    a decisive factor for panoptic segmentation; 
    it allows to explicitly model the learning dynamics, 
    and reconcile the learning of multiple objectives (up to ten in our experiments);
    with an end-to-end architecture, 
    it generalizes to different datasets without the need of re-tuning hyperparameters or 
    re-adjusting the training process laboriously. 
    Our Ada-Segment brings 2.7\% panoptic quality (PQ) improvement on COCO \textit{val} split 
    from the vanilla baseline, 
    achieving the state-of-the-art $48.5\%$ PQ on COCO \textit{test-dev} split 
    and 32.9\% PQ on ADE20K dataset.
    The extensive ablation studies reveal the ever-changing dynamics 
    throughout the training process, 
    necessitating the incorporation of an automated and adaptive learning strategy as presented 
    in this paper. 
\end{abstract}

\begin{figure}[t]
    \centering
    \includegraphics[width=1.0\linewidth]{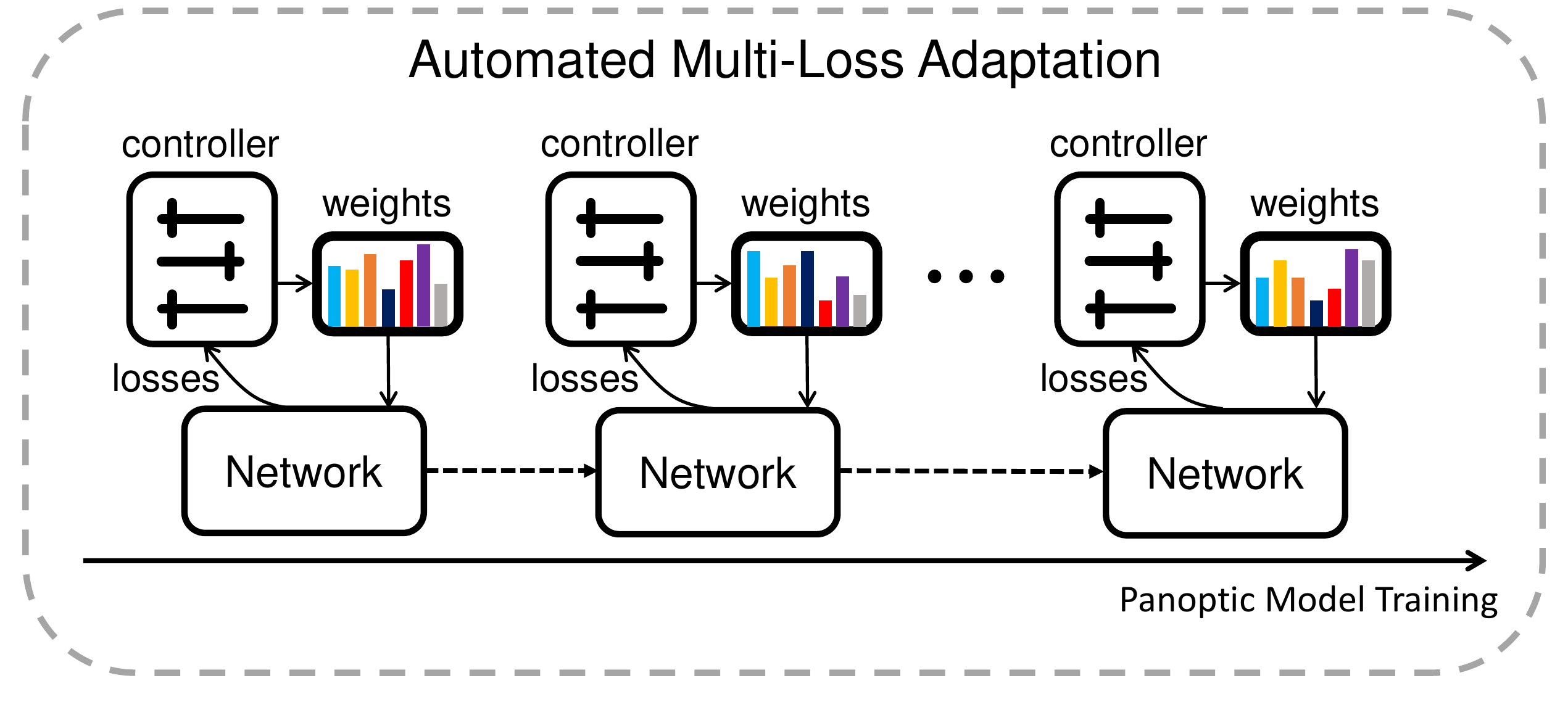}
    \caption{Our Ada-Segment aims at automatically adjusting the weights of multiple objectives 
    in panoptic segmentation during training for achieving the balanced learning dynamics, 
    in contrast to existing works that rely on carefully hand-tuned weights after tediously re-training 
    multiple times. 
    It adjusts the training loss every training epochs on the fly via a weight controller 
    within a single training procedure. 
    Ada-Segment can achieve remarkably better results than results without loss tuning 
    and also perform significant better than other hyperparameter tuning methods. }
    \label{intro2}
 \end{figure}
 \section{Introduction}
 Capitalized on advances from traditional semantic segmentation and instance segmentation, 
 the vision community  recently steps forward to resolve a more challenging task, 
 panoptic segmentation~\cite{panoptic_segmentation}, 
 which targets at simultaneously segmenting both foreground instance \textit{things} 
 (\textit{e.g.}, person, car and dog) and background semantic \textit{stuff} (\textit{e.g.}, sky, river and sea), 
 achieving a more unified understanding of images. 
 Hence, panoptic segmentation is usually formulated as a multi-objective problem 
 for jointly optimizing for semantic and instance segmentation. 
 To solve this problem, many existing methods~\cite{panopticFPNkirillov2019,upsnetxiong2019,seamlessporzi2019,spatialflowchen2019} 
 have designed multi-branched network architectures, 
 while each branch mapping to an instance or semantic segmentation objective 
 and resulted in many (up to ten in our experiments) 
 individual losses that need to be reconciled during training.
 
 Various network modules or fusing strategies~\cite{panopticFPNkirillov2019,upsnetxiong2019,oanetLiu2019,lazarow2020learningocfusion} 
 have been proposed to deal with the consistency of learning 
 or predictions from instance segmentation and semantic segmentation branches. 
 As reported in many literatures~\cite{panopticFPNkirillov2019,upsnetxiong2019},
 the performance of a panoptic segmentation architecture exhibits extreme sensitivity and variability 

 with respect to the multi-objective loss weights.
 Therefore, previous works relied on exhaustive hyper-parameter search over such weights. 
 For example, Panoptic FPN~\cite{panopticFPNkirillov2019} uses a grid search to find better loss weights 
 on two datasets; UPSNet~\cite{upsnetxiong2019} carefully investigates the weighting scheme 
 of loss functions. In our experiments, different loss weights may yield 2\% performance difference 
 (measured in PQ).
 It is thus hard to disentangle the advantages of an improved method from a better hyperparameter 
 setting. 
 
 Moreover, previous works only apply static loss weights throughout the training, 
 skipping chances for the appropriate adaptation to the dynamically-varying convergence 
 behaviors, as observed in our experiments. 
 Finally, hand-tuned loss weights, whenever changing to a different dataset, 
 must be carefully re-tuned, 
 prohibiting the generalization across different data distributions.
 
 To address these limitations, 
 we present Ada-Segement, an efficient automated multi-loss adaptation framework 
 to dynamically adjust the loss weight with respect to each sub-objective, 
 seeking an improved optimization schedule during training.
 In Figure~\ref{intro2}, Ada-Segment introduces an end-to-end weight controller 
 to automatically generate loss weights to adjust model's training loss.
 Departing from fixing a group of static weights during training, 
 Ada-Segment adjusts the loss weights based on the training conditions 
 with the weight controller.
 Specifically, 
 the weight controller is firstly trained with several models training in parallel.
 It gathers training information from all models 
 after a few training iterations (\textit{e.g., an epoch})
 and produces new weights for training. 
 Besides, we find the trained controller is of the capability
 to capture the ever-changing 
 training dynamics so that we can directly re-use it to automatically adjust training loss when training models
 on different datasets, training schedules and backbone networks. 

 Our main contributions are summarized as follows:
 (1) We propose Ada-Segment as a framework to automatically balance the multiple objectives 
 in panotpic segmentation, bypassing tedious manual tuning of loss combination weights.
 (2) We introduce a novel weight controller within the multi-loss adaptation strategy 
 that can capture the learning dynamics to adjust the loss weights during training, 
 which is of the capability to transfer between different backbone network, training schedule and 
 datasets. 
 3) We empirically demonstrate the significance of the convergence dynamics in 
 panoptic segmentation model training. 
 (4) Ada-Segment significantly outperforms previous state-of-the-art
 methods on COCO \textit{test-dev} dataset with 48.5\% PQ 
 and ADE20K with 32.9\% PQ, 
 and brings 2.7\% panoptic quality (PQ) improvement on COCO \textit{val} split.
 Extensive ablation studies verify the importance 
 of automated multi-loss adaptation and the generalizability of the framework.
 
\section{Related Work}

 \textbf{Panoptic Segmentation. } 
 The recently proposed panoptic segmentation task~\cite{panoptic_segmentation} 
 departs from traditional multi-task problem~\cite{imageParsingtu2005,farhadi2009describing} 
 by introducing a unified task with meticulously designed task metrics, 
 which requires algorithms to output unified results in a single model. 
 Several works~\cite{panopticFPNkirillov2019,seamlessporzi2019} approach this problem via combining well-developed instance segmentation 
 models~\cite{he2017mask} with a semantic segmentation decoder, 
 and fusing the results together~\cite{oanetLiu2019,panopticFPNkirillov2019,seamlessporzi2019}. 
 Besides, some works improve the interaction between sub-tasks through reasoning modules~\cite{wu2020bgr}, 
 attention mechanisms~\cite{chen2020banet,AunetLi_2019_},
  unified head~\cite{upsnetxiong2019}, automated neural architecture search~\cite{wu2020autopanoptic} or even deploy a bottom-up approach~\cite{cheng2019panopticdeeplab}. 
 However, none of them have developed an effective or automated way to alleviate the imbalance caused by multiple subtasks; 
 Instead, they spend a numerous amount of time trying to adjust the loss weights by hand. 
 In this work, we aim at tackling this problem with an automated adaptation strategy. 
 
 \noindent
 \textbf{Multi-objective Learning}. 
 Panoptic Segmentation is a unified computer vision task derived from multi-objective learning. 
 However, when lacking systematic treatments, using a single model to handle multiple tasks 
 may downgrade the performance~\cite{kokkinos2017ubernet} of the target task. 
 Some optimization technics are proposed by previous works~\cite{kendall2018uncertainty,chen2017gradnorm}, 
 however, they are still unstable and even lead to divergence.
 Accordingly, for panotpic segmentation, some 
 works~\cite{upsnetxiong2019,panopticFPNkirillov2019,oanetLiu2019} 
 try to manually find weights to adjust the training process and balance among subtasks, expecting to get higher performance, 
 which, however, is time-intensive and cannot generalize across datasets -- a problem that this paper tries to address.
 
 \noindent
 \textbf{Adaptive Learning}. Adaptive learning is a widely researched topic. 
 Curriculum Learning~\cite{bengio2009curriculum,lin2018focal,ren2018autoreweigthing} 
 proposes to gradually increase the difficulty of training samples during training. Along this line, 
 closest to ours is AutoLoss~\cite{xu2018autoloss}, 
 which uses reinforcement learning to learn update schedules in alternate optimization problems.

 \noindent
 \textbf{Hyperparameter Tuning}. Hyperparameter tuning is of a long history~\cite{feurer2019HPOsuvey}. 
 Traditional sample-based methods like grid or random search are computationally costly. 
 PBT~\cite{jaderberg2017pbt} partly relieves this problem by tuning hyperparameters during training 
 but cannot achieve satisfactory results. 
 Bayes Optimization-based methods like GP-BO~\cite{snoek2012GPBO} and SMAC~\cite{hutter2011SMAC} 
 utilize Bayes Optimization to achieve better tuning performance but 
 are also computationally inefficient.
 BOHB~\cite{falkner2018bohb} proposes a more efficient BO-based method that combines Bayes Optimization
 with Hyperband~\cite{li2017hyperband} to accelerate to the searching process. 
 However, it is not applicable for searching hyperparameters dynamically. 
 Gradient-based methods~\cite{zeiler2012adadelta,baydin2017HDsgd,pedregosa2016AGhpo} benefit the tuning of some hyperparameters 
 like learning rate during training but are hard to generalize well to the scenario 
 of multi-objective weighting. 
 RL-based methods like~\cite{huang2019addressing} designs specific search space for parameters within
 classification and metric learning loss functions, which cannot generalize to the scenario of 
 multi-objective weighting; AM-LFS~\cite{li2019AMlfs} directly tunes specific hyperparameters with greedy strategy, 
 which ignores the training conditions and results in a sub-optimal solution.

 \section{Automated Online Multi-loss Adaptation}
 \label{rec:Dm-ada}
 
 \begin{figure}[t]
   \centering
   \includegraphics[width=1.0\linewidth]{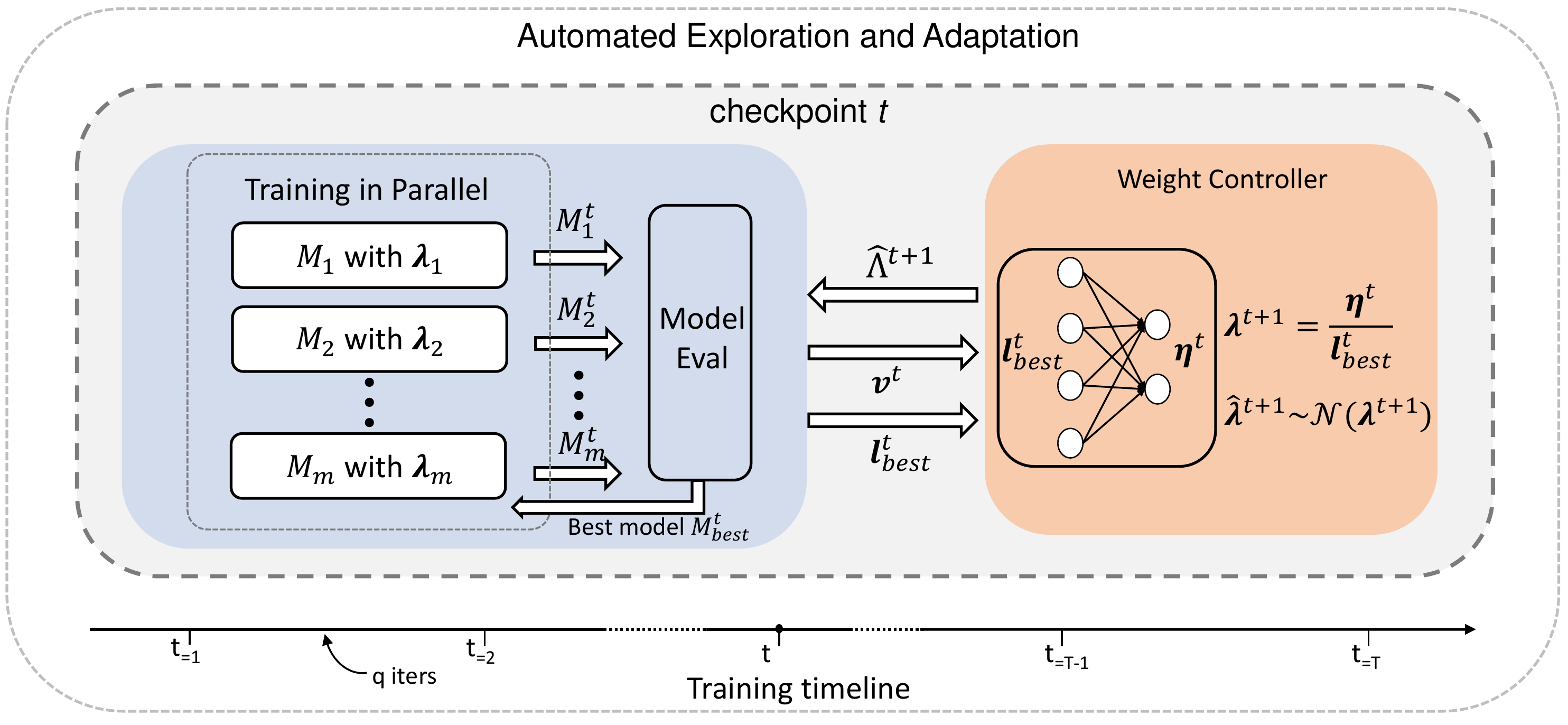}
   \caption{
   Illustration of the Automated Exploration and Adaptation in our Ada-Segment framework. 
   It views the whole training process as a series of checkpoints ( $T$ checkpoints in total).
   The weight controller is trained interactively with 
   $m$ models trained in parallel and evaluated at each checkpoint. }
   \label{Overview}
 \end{figure}

 \subsection{Overview}
 The common architectures~\cite{panopticFPNkirillov2019,upsnetxiong2019} tackle panoptic segmentation using 
 a multi-objective model with several additional losses 
 (\textit{e.g.}, losses from box head, segmentation head etc.).
 Given the loss vector $\boldsymbol{l} \in \mathbb{R}^n$ and their corresponding loss weight vector 
 $\boldsymbol{\lambda} \in \mathbb{R}^n$ where $n$ is the number of training losses,
 we define the weighted loss of our panoptic segmentation framework as $L=\sum_{i=1}^n\lambda_{i}l_{i}$.

 As stated previously, multi-loss weighting is essential but difficult in panotpic segmentation.
 Inspired by the adaptive learning and automated machine learning methods,
 the goal of our automated multi-loss adaptation (named Ada-Segment)
 framework is to automatically adjust $\boldsymbol{\lambda}$ during training via a controller. 
 As Figure~\ref{Overview} shows, the controller is jointly trained 
 with $m$ models (\textit{i.e.}, panotpic segmentation networks) $\{M_1, M_2, ..., M_m \}$ 
 training in parallel on a proxy training dataset (training set in short).
 After that, it can be used for controlling single model training at anytime.

 \subsection{Weight Controller}
 \label{sec:w-gen}
 The weight controller is proposed to capture the training dynamics 
 and automatically adjust loss weights during training. 
 At the $t^{th}$ checkpoint, suppose the network produces the loss vector $\boldsymbol{l}^t$, 
 and we want to find a weight vector $\boldsymbol{\lambda}^{t+1}$ to adjust the loss value 
 so that the weighted loss can guide the network towards better optimization. 
 Directly determine $\boldsymbol{\lambda}$ is impossible since we have no prior about it. 
 However, since we know that the weighted loss vector is formulated as 
 $\boldsymbol{\eta}^t=\boldsymbol{\lambda}^{t+1}\odot \boldsymbol{l}^t$
 where $\odot$ is the element-wise multiplication. 
 We introduce the weight controller to estimate the weighted loss vector as the 
 transformation of the current loss vector as $\boldsymbol{\eta}^t=\pi(\boldsymbol{l}^t; \theta),$
 where $\theta$ is the learnable parameter in the weight controller $\pi$. 
 Therefore, we can obtain the weight vector $\boldsymbol{\lambda}^{t+1}$ by 
 \begin{equation}\boldsymbol{\lambda}^{t+1}=\frac{\boldsymbol{\eta}^t}{\boldsymbol{l}^t}=
 \frac{\pi(\boldsymbol{l}^t; \theta)}{\boldsymbol{l}^t}.\label{eq:pred}\end{equation}

 \noindent
 \textbf{Weight Controller Optimization.}
 In this work, we use a policy network as the weight controller to predict the estimated weighted loss 
 based on the loss at each time checkpoint. 
 Since the loss weights directly determine the training loss, it may get into a dilemma if we use 
 training loss to optimize the policy network through back-propagation. 
 Some optimization technics are proposed by previous works~\cite{kendall2018uncertainty,chen2017gradnorm}, 
 however, they are still unstable and even lead to divergence when 
 directly using training gradients to optimize weights on complex tasks like panoptic segmentation.
 Therefore, we optimize the policy network towards the evaluation metric 
 (\textit{i.e.}, PQ in panoptic segmentation) 
 via the validation set through REINFORCE~\cite{williams1992REINFORCE}.

 \textbf{State Space}: 
 During training, at any time checkpoint $t$,
 the policy network takes the loss vector $\boldsymbol{l}^t$ to represent the optimization state of the model,
 and outputs the estimated weighted loss 
 $\boldsymbol{\eta}^t$ to calculate $\boldsymbol{\lambda^{t+1}}$. 
 
 \textbf{Action Space}: Exploration is of great importance to find improved solution via an 
 action space design. We sample loss weight candidates 
 $\hat\Lambda^{t+1}=\{\boldsymbol{\hat\lambda}^{t+1}_1,\boldsymbol{\hat\lambda}^{t+1}_2,...,
 \boldsymbol{\hat\lambda}^{t+1}_m\}$ to train $m$ models
 from a Normal distribution 
 \begin{equation}\boldsymbol{\hat\lambda}^{t+1}\sim\mathcal{N}(\boldsymbol{\lambda^{t+1}},\sigma),\label{eq:sample}\end{equation}
 in which $\boldsymbol{\lambda^{t+1}}$ is uesd as the mean value and 
 $\sigma$ is the sampling standard deviation.

 \textbf{Rewards}: Reward function $r(\cdot)$ measures the quality of the generated actions. 
 Intuitively, after applying the actions, we can use the models' performances (PQ) as the 
 policy rewards to guide the training of the policy network. We refer this as the local reward function
 \begin{equation}r_{local}(\boldsymbol{v}^t)=\frac{\boldsymbol{v}^t-mean(\boldsymbol{v}^t)}{std(\boldsymbol{v}^t)},\end{equation}
 which normalizes
 the validation performances $\boldsymbol{v}^t\in \mathbb{R}^m$ at checkpoint $t$ 
 to zero mean and unit variance as rewards.
 
 Furthermore, we include the relative improvement from the previous checkpoint as the policy rewards:
 \begin{equation}r_{imp}(\boldsymbol{v}^t_{imp})=\frac{\boldsymbol{v}^{t}_{imp}}{std(\boldsymbol{v}^t_{imp})},\end{equation}
 which calculates the normalized absolute improvement from the previous checkpoint to introduce long-range influences
 where $\boldsymbol{v}^t_{imp} = \boldsymbol{v}^t-v^{t-1}_{best}.$
 This is non-trivial because only using the differences between temporal samples would
  overlook the training dynamics between checkpoints.

 Therefore, the overall rewards are calculated as  
 \begin{equation}r(\boldsymbol{v}^t)=\frac{t}{T}(r_{local}(\boldsymbol{v}^t)
  +r_{imp}(\boldsymbol{v}^t_{imp})),\label{eq:reward}\end{equation}
 where the scale factor $\frac{t}{T}$ controls the magnitude of the 
 overall reward according to the training process 
 since the early training stages are less important with more randomness.

 \textbf{Parameter Updates:}
 Given the sample rewards, the parameter $\theta$ of the policy network $\pi$ is updated by 
 the gradients 
 \begin{equation}\nabla_{\theta}R^t(\theta)=\frac{1}{m}\sum_{j=1}^{m}
 {r^t_j(\boldsymbol{v}^t)\nabla_{\theta}\log{s(\boldsymbol{\hat\lambda}^{t+1}_j;
 \frac{\pi(\boldsymbol{l}^t; \theta)}{\boldsymbol{l}^t}, \sigma)}},\label{eq:reinforce}\end{equation}
 where $s(\cdot;\mu,\sigma)$ is the probability density function of the Normal distribution.
 It is worth noting that we only consider the situation that all loss values are nonnegative,
 which is the common scenario in panoptic segmentation. 
 Therefore, when sampling from the Normal distribution, 
 samples contain negative values would be given $-1$ as the reward directly to increase the training stability. 

 \subsection{Ada-Segment Algorithm} 
 \label{sec:ada-segment-alg}
 In this section, we introduce how the overall Ada-Segment framework works in detail. 

 \noindent
 \textbf{Initial State}. 
 Since the policy network requires training losses as input. At the very beginning, 
 one pseudo training epoch is performed with all loss weights equal to 1 before the exact training schedule
 to obtain the initial loss $\boldsymbol{l}^1$. 

 \noindent
 \textbf{Policy Initilaization} The initial policy is crucial for training stability. 
 All layers in the policy network are randomly initialized 
 by the Normal distribution with mean value equal to $1/n_c$ instead of $0$ to avoid the loss weights 
 to be non-positive at the beginning (except the bias parameters is initialized to 0), 
 where $n_c$ is the number of input channels of each layer.

 \noindent
 \textbf{Automated Exploration and Adaptation}.
 In this phase, the controller and $m$ models are jointly trained as shown in Figure~\ref{Overview}. 
 Between two checkpoints, we train models in parallel with 
 separate loss weights generated by the weight controller and track the training losses. 
 At each checkpoint, we evaluate all models on the validation set,
 obtaining the model performances $\boldsymbol{v}^t$ (\textit{i.e.}, PQ value on validation for panoptic segmentation)
 to calculate rewards following Equatin~\ref{eq:reward}.
 The policy network is  
 updated by gradients descent via Equation~\ref{eq:reinforce}. 
 To continue training, 
 we broadcast the best-performed model to all other models,
 thus we can use $l^t_{best}$ as the training condition at each time checkpoint. 

 \noindent
 \textbf{Policy Transfer}.
 We can get the best-performed model $M^T_{best}$ after the exploration phase. 
 However, we take it a step further and propose the policy transfer. 
 Our framework, once finished one exploration 
 process, also produces the trained controller that 
 gives us the chance to re-use it at anytime,
 saving the time and computational cost when the dataset or training schedule changes.
 
 Along with the training of the model, the policy network is changing accordingly, causing the 
 policy $\pi^T$ may favor the latest training condition but partly forget former loss situations. 
 Besides, earlier states of policy may suffer from the under-fitting problem with few update iterations. 
 Therefore, to make full use of the controller, 
 we proposed to combine all states of the policy network during exploration phase 
 \textit{i.e., }$\pi^1$ to $\pi^T$, via a weighted
 \textit{policy ensemble} strategy.

 When training a model $M_p$ for $E$ epochs and adjusting loss weights every training epoch, 
 with $E$ is not necessary to be equal to $T$. 
 By calculating the distance between a training epoch $e$ and corresponding update checkpoint $t$, 
 we can assign a weight to each policy state at different training epochs controlled by 
 a discount factor $\gamma=0.9$. Therefore, we have 
 
 \begin{equation}
   \boldsymbol{\lambda}^{e+1}=\frac{1}{Z}\sum^T_t{\gamma^{\lvert \frac{e\times T}{E}-t \rvert}\frac{\pi^t(\boldsymbol{l}^e;\theta)}
   {\boldsymbol{l}^e}},
  \label{eq:policy_ensemble}
 \end{equation}
 where $\frac{T}{E}$ align the training procedure with the number of exploration checkpoints and
 $Z=\sum^T_t{\gamma^{\lvert \frac{e\times T}{E}-t \rvert}}$ is the normalizing factor.
 By combining policy states at different stages, the training dynamics
 captured by the policy network can be preserved to a large extend.
 To sum up, the paradigm of our Ada-Segment is summarized in Algorithm~\ref{alg1}.  
  
\begin{algorithm}
  $\textbf{Input}$: Iterations between two checkpoints $q$, Initial Loss State $\boldsymbol{l}^1$, 
  Number of checkpoints $T$, Number of training epochs $E$
  
  \quad Initialize $m$ models $\{M_1,M_2,...M_m\}$

  \quad Initialize policy network $\pi$




  \quad $\boldsymbol{l}_{best}^1 \leftarrow \boldsymbol{l}^1$

  \quad $\textbf{for}$ $t \leftarrow 1$ to $T$, $\textbf{do}$
  
  \quad \quad Generate $\boldsymbol{\lambda}^{t+1}$ by Equation~\ref{eq:pred} 
  with $\pi$ and $\boldsymbol{l}_{best}^t$

  \quad \quad Sample $m$ candidates by 
  Equation~\ref{eq:sample} with $\boldsymbol{\lambda}^{t+1}$
  
  \quad \quad Train all models for $q$ iterations with $\hat\Lambda^{t+1}$

  \quad \quad Collect model performances $\boldsymbol{v}^t$ and save $\boldsymbol{l}_{best}^t$

  \quad \quad Obtain policy rewards by Equation~\ref{eq:reward}
  
  \quad \quad Update $\theta$ in $\pi$ via Equation~\ref{eq:reinforce}

  \quad \quad Save $\pi^t \leftarrow \pi$
  
  \quad \quad Update all models 
  with
  $M_{best}^{t}$


  
  






  \quad $\textbf{end for}$

  \quad Initialize a model $M_{p}$

  \quad $\textbf{for}$ $e \leftarrow 1$ to $E$, $\textbf{do}$

  \quad \quad Generate $\boldsymbol{\lambda}^{e+1}$ by Equation~\ref{eq:policy_ensemble} with $\boldsymbol{l}^e$

  \quad \quad Train model $M_{p}$ with $\boldsymbol{\lambda}^{e+1}$

  \quad $\textbf{end for}$
  
  $\textbf{return}$ $M^p$
  \caption{The Ada-Segment framework. }
  \label{alg1}
\end{algorithm}

 \begin{figure}[t]
  \centering
  \includegraphics[width=1.0\linewidth]{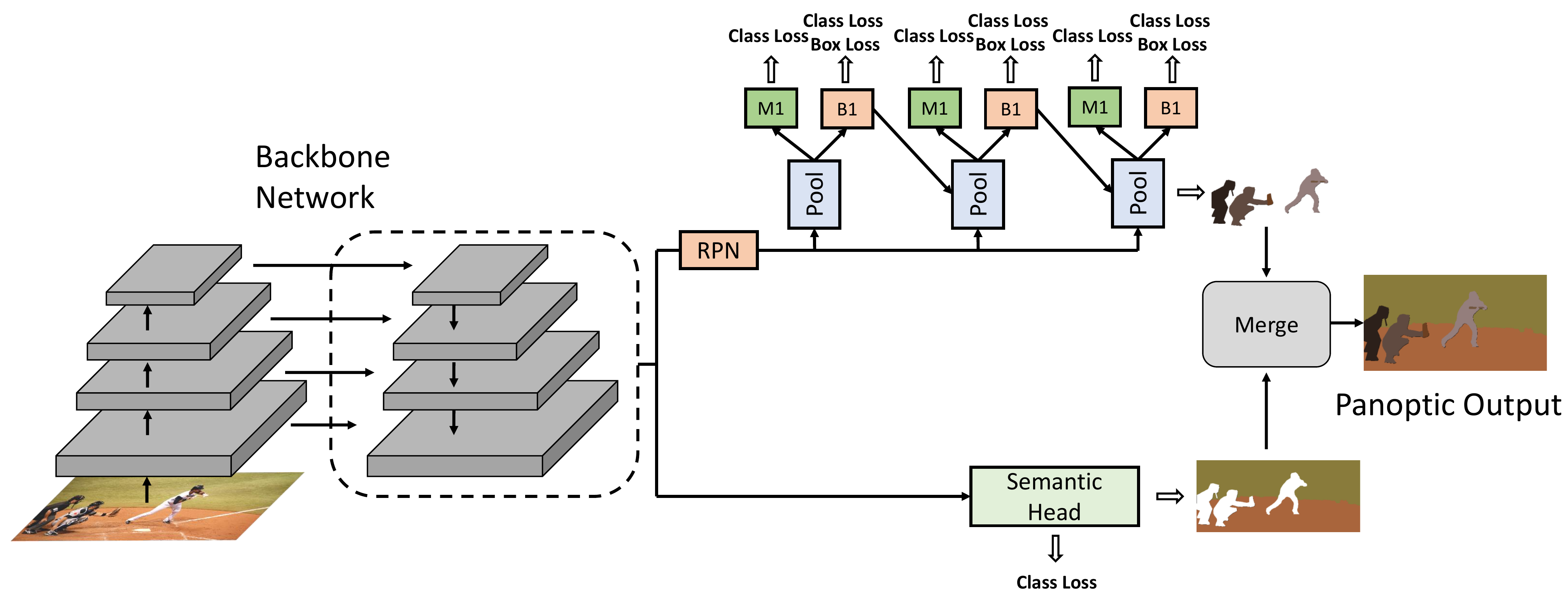}
  \caption{The network structure of our baseline. For detection branch, we use the three-stage  
  Cascade R-CNN~\cite{cai2017cascade}, which contains three pairs of losses. 
  The semantic segmentation branch is a simple SemanticFPN in~\cite{panopticFPNkirillov2019}. 
  The overall architecture has ten losses to be jointly optimized. }
  \label{fig:network_structure}
\end{figure}

 \subsection{Network Structure}
 \label{subsec:Network-Structure.}
 
 Figure~\ref{fig:network_structure} shows the network structure used in our experiments,
 which extends Cascade Mask R-CNN~\cite{cai2017cascade} with a semantic segmentation branch.

 \noindent
 \textbf{Backbone Networks.} 
 We use a ResNet~\cite{resnetHe2015}
 with a Feature Pyramid Network (FPN)~\cite{fpnlin2017}. 
 and add deformable convolution~\cite{dai2017deformable} 
 in stage $3\sim5$ of the backbone networks.
 
 \noindent
 \textbf{Instance Segmentation Head.} 
 We use three cascaded detection heads after region proposal network 
 in our network. 
 To get the instance segmentation results, each stage of head outputs bounding box regression, 
 classification and mask results for objects in an image. 
 
 \noindent
 \textbf{Semantic Segmentation Head.} 
 We use a semantic segmentation head following Panotic FPN~\cite{panopticFPNkirillov2019}.
 It takes the FPN features as inputs and uses 1x1 convolution and bi-linear upsample 
 function to gradually upsample each FPN feature to 1/4 of input image size.
 All upsampled features are summed up and transformed into the final segmentation map 
 by a 1x1 convolution.

 \section{Experiments}
 \label{sec:exp}

 \subsection{Datasets and Evaluation}
 \noindent
 \textbf{COCO}. Following the competition setting in 2019 Microsoft COCO panoptic segmentation, 
 which consists of 133 classes with 80 things classes and 53 stuff classes. We only use \textit{train2017} 
 split with approximately 118k images for training and report the results on \textit{val} split 
 with 5k images. We also report our results on COCO \textit{test-dev} split for comparison 
 with other state-of-the-art methods.
 
 \noindent
 \textbf{ADE20K}. ADE20K is a challenging dataset with densely labeled 22k images, 
 with 100 things classes and 50 stuff classes. 
 It contains heavier occlusions, more tiny objects and class ambiguities than COCO,
  and thus more challenging.
 
 \noindent
 \textbf{Proxy Dataset Setting}. 
 In the practice of automated machine learning,
 for evaluation during training, it is necessary to construct a proxy validation
 dataset instead of using the original validation set. 
 For COCO, we randomly sample 10k images from the 118k training set 
 for validation and the rest part as the \textit{proxy} training set. 
 For ADE20K, we train on 20k training images in which 2k images are
  randomly sampled images for validation during training.

 \noindent
 \textbf{Evaluation Setup}.  We follow the panoptic segmentation evaluation
 metrics proposed in~\cite{panoptic_segmentation} to evaluate our
 models in terms of panoptic quality (PQ), segmentation quality (SQ)
 and recognition quality (RQ). Note that PQ is the weighted sum of PQ$^{th}$ and PQ$^{st}$
 for things and stuff classes respectively. We report these two metrics for demonstrating 
 the effectiveness of our method.

 \begin{table}[t]
  \begin{centering}
  \tabcolsep 0.03in\renewcommand{\arraystretch}{1.2}{\footnotesize{}}%
  \begin{tabular}{c|c|ccc}
  \hline 
  Method & training set & PQ & PQ$^{th}$ & PQ$^{st}$ \tabularnewline
  \hline 
  Baseline & COCO$_p$ & 40.7 & 47.1 & 31.0 \tabularnewline
  Baseline-G & COCO$_p$ & 41.7$^{+1.0}$ & 48.7$^{+1.6}$ & 31.1$^{+0.1}$ \tabularnewline
  Baseline-P & COCO$_p$ & 41.1$^{+0.4}$ & 48.8$^{+1.7}$ & 29.4$^{-1.6}$ \tabularnewline
  \hline 
  Ada-Segment-A & COCO$_p$ & 42.6$^{+1.9}$ & 49.5$^{+2.4}$ & 32.1$^{+1.1}$ \tabularnewline
  Ada-Segment & COCO$_p$ & 43.2$^{+2.5}$ & 50.5$^{+3.4}$ & 32.3$^{+1.3}$ \tabularnewline
  \hline 
  \hline 
  Baseline & COCO & 41.0 & 47.2 & 31.5 \tabularnewline
  Baseline-G & COCO & 42.1$^{+1.1}$ & 49.0$^{+1.8}$ & 31.6$^{+0.1}$ \tabularnewline
  \hline 
  Ada-Segment & COCO & 43.7$^{+2.7}$ & 51.2$^{+4.0}$ & 32.5$^{+1.0}$ \tabularnewline
  \hline 
  \end{tabular}{\footnotesize\par}
  \par\end{centering}
  \caption{\label{tab:Ablation-Studies-on-coco}Comparison with different baselines on COCO \textit{val} split.
  All models are trained on the proxy training set.   
  \textit{Baseline-G}: using coarse grid search to optimize loss weights with multiple runs.
  \textit{Baseline-P}: applying a PBT-like~\cite{jaderberg2017pbt} framework to tune 
  loss weights during training, which can be viewed as our method without the weight controller. $COCO_p$ represents the proxy dataset.}
  \end{table}

  \begin{table}[t]
    \begin{centering}
    \tabcolsep 0.03in\renewcommand{\arraystretch}{1.2}{\footnotesize{}}%
    \begin{tabular}{c|c|ccc}
    \hline 
    Method & Weighting Type & PQ & PQ$^{th}$ & PQ$^{st}$ \tabularnewline
    \hline 
    Baseline & static & 41.0 & 47.2 & 31.5 \tabularnewline
    Baseline-G & static & 42.1 & 49.0 & 31.6 \tabularnewline
    Final & static & 42.2 & 50.0 & 30.4 \tabularnewline
    Pred & static & 42.6 & 50.3 & 31.0 \tabularnewline
    Single-dy & dynamic & 43.1 & 50.1 & 32.4 \tabularnewline
    Comb-dy & dynamic & 43.7 & 51.2 & 32.5 \tabularnewline
    \hline 
    \end{tabular}{\footnotesize\par}
    \par\end{centering}
    \caption{\label{tab:ana-dy} Comparison of different static weighting strategy and different 
    dynamical adaptation strategy on COCO.}
  \end{table}

  \begin{table*}[t]
    \begin{centering}
    \tabcolsep 0.05in\renewcommand{\arraystretch}{1.2}{\footnotesize{}}%
    \begin{tabular}{c|c|c|c|c|c|ccc}
    \hline
    Train Arch. & Trans. Arch & Train Data & Trans. Data & Train. Sche. & Tran. Sche. & PQ & PQ$^{th}$ & PQ$^{st}$\tabularnewline
    \hline 
    \hline
    R-50 & R-50 & ADE20K$_p$ & COCO & 1x & 1x & 42.7 & 49.0 & 33.1 \tabularnewline
    R-50 & R-50 & COCO$_p$ & COCO & 1x & 1x & 43.7 & 51.2 & 32.5 \tabularnewline
    R-50 & R-50 & COCO$_p$ & ADE20K & 1x & 1x & 32.0 & 34.3 & 27.4 \tabularnewline
    R-50 & R-50 & ADE20K$_p$ & ADE20K & 1x & 1x & 32.9 & 35.6 & 27.9\tabularnewline
    \hline 
    R-50 & R-101 & COCO$_{p}$ & COCO & 1x & 1x & 45.1 & 52.7 & 33.6 \tabularnewline
    R-101 & R-101 & COCO$_{p}$ & COCO & 1x & 1x & 45.2 & 52.2 & 34.7 \tabularnewline
    \hline
    R-50 & R-50 & COCO$_{p}$ & COCO & 1x & 2x & 44.3 & 51.3 & 33.7 \tabularnewline
    R-50 & R-50 & COCO$_{p}$ & COCO & 2x & 2x & 44.4 & 50.9 & 34.5 \tabularnewline
    \hline
    \end{tabular}{\footnotesize\par}
    \par\end{centering}
    \caption{\label{tab:transfer} Transferability of the policy network across 
    different backbones, training schedules and datasets.
    $D_p$ means searching and training on the proxy dataset.
    1x: 12 epochs on COCO or 24 epochs on ADE20K; 2x: training for 24 epochs on COCO.}
  \end{table*}

  \begin{table}[t]
    \begin{centering}
    \tabcolsep 0.05in\renewcommand{\arraystretch}{1.2}{\footnotesize{}}%
    \begin{tabular}{c|ccc}
    \hline 
    Method & PQ & PQ$^{th}$ & PQ$^{st}$\tabularnewline
    \hline 
    Panoptic FPN$^{\dagger}$ & 38.1 & 43.8 & 29.4\tabularnewline
    Panoptic FPN$^{\ddagger}$ & 39.0$^{+0.9}$ & 46.1$^{+2.3}$ & 28.3$^{-1.1}$\tabularnewline
    Panoptic FPN$^{\star}$ & 39.0$^{+0.9}$ & 45.9$^{+2.1}$ & 28.7$^{-0.7}$\tabularnewline
    w Ada-Segment & 39.9$^{+1.8}$ & 46.6$^{+2.8}$ & 29.7$^{+0.3}$\tabularnewline
    \hline 
    \end{tabular}{\footnotesize\par}
    \par\end{centering}
    \caption{\label{tab:Evaluation-on-panopticfpn}Multi-loss adaptation for Panoptic FPN~\cite{panopticFPNkirillov2019} on COCO, 
    ${\dagger}$: our re-implementation with all loss weights equal to 1; $\ddagger$: our re-implementation
    with loss weights used in the paper; $\star$: results reported in the paper.}
  \end{table}

 \subsection{Implementation Details}
 \noindent
 \textbf{Model Training}. We follow the commonly used hyper-parameter settings as stated in 
 Panoptic FPN~\cite{panopticFPNkirillov2019} within our experiments. 
 We set the initial learning rate as 0.02
 and weight decay as 0.0001 with stochastic gradient descent
 (SGD) for all experiments. 
 To provide more stable information to train the policy network, we decrease learning
 rate with cosine policy, which is more smooth than decrease the learning
 rate at specific iterations. We initialize the backbone network with ImageNet
 pretrained model while the remaining parameters are initialized following~\cite{he2015init}. 
 For each model, we train totally 12 epochs (so called 1x setting) for COCO 
 and 24 epochs for ADE20K 
 on 8 GPUs with 2 images per GPU using PyTorch~\cite{paszke2017automatic}.

 \noindent
 \textbf{Automated Multi-loss Adaptation}. 
 During joint training, we deploy $m=8$ models, where each model contains $n=10$ losses:
 three pairs of instance detection losses (bounding box loss, classification
 loss and mask loss) and a single semantic segmentation loss.
 In the weight controller, we set the sampling standard deviation $\sigma=0.2$,
 and we use three-layer MLP with hidden 
 layer size 16 as the policy network. We use Adam~\cite{kingma2014adam} optimizer with learning rate 
 $5e^{-2}$ and weight decay $5e^{-4}$ to optimize the policy network. 
 For the overall Ada-Segement framework, we simply train 1 Epoch between two checkpoints for both COCO 
 and ADE20K datasets. When adjusting loss weight, we re-scale the learning rate of the $i^{th}$ head by $\frac{1}{\lambda_{i}}$ 
 for $i=1,2,...,n$ to ensure the head networks to be fully trained so that the loss weights only
 influence the shared backbone and the detection losses are averaged among three cascade stages.  
 
 \noindent
 \textbf{Inference}. The panoptic results are obtained 
 in the way proposed in~\cite{panoptic_segmentation}.
 Specifically, after merging instance masks on the non-overlap canvas, the remaining pixels
 are assigned according to semantic segmentation results (with areas less than 4096 being ignored). 
 
 \noindent
 \textbf{Baseline Setup}. The baseline method means 
 setting equaling weights (with value 1) during training. 
 As to be shown in the following sections, although a strong baseline network is used, 
 it only produces unsatisfactory results, which is inhibited by the improper weight setting.
 Without particular notions, we report results with ResNet-50 backbone.

\begin{table*}
  \begin{centering}
  \tabcolsep 0.1in{\scriptsize{}}%
  \begin{tabular}{c|c|ccc|cc}
  \hline 
  Methods & backbone & PQ & PQ$^{th}$ & PQ$^{st}$ & SQ & RQ\tabularnewline
  \hline 
  \hline 
  Panoptic FPN~\cite{panopticFPNkirillov2019} & ResNet-101-FPN & 40.9 & 48.3 & 29.7 & - & -\tabularnewline
  OANet~\cite{oanetLiu2019} & ResNet-101-FPN & 41.3 & 50.4 & 27.7 & - & -\tabularnewline
  AUNet~\cite{AunetLi_2019_} & ResNeXt-152-FPN-D & 46.5 & \underline{55.8} & 32.5 & 81.0 & 56.1\tabularnewline
  UPSNet~\cite{upsnetxiong2019} & ResNet-101-FPN-D & 46.6 & 53.2 & 36.7 & 80.5 & 56.9\tabularnewline
  OCFusion~\cite{lazarow2020learningocfusion} & ResNeXt-101-FPN-D & 46.7 & 54.0 & 35.7 & - & -\tabularnewline 
  SpatialFlow~\cite{spatialflowchen2019} & ResNet-101-FPN-D & 47.3 & 53.5 & \underline{37.9} & \underline{81.8} & 56.9\tabularnewline
  BANet~\cite{chen2020banet} & ResNet-101-FPN-D & 47.3 & 54.9 & 35.9 & 80.8 & 57.5\tabularnewline
  SOGNet\cite{yang2019sognet} & ResNet-101-FPN-D & 47.8 & - & - & 80.7 & 57.6\tabularnewline
  \hline 
  Ours & ResNet-101-FPN-D & \textbf{48.5} & \underline{55.7} & \underline{37.6} & \underline{81.8} & \textbf{58.2} \tabularnewline
  \hline 
  \end{tabular}{\scriptsize\par}
  \par\end{centering}
  
  \caption{\label{tab:Results on COCO test-dev} Results on COCO \textit{test-dev} split. 
  In the table, '-D' represents methods using deformable convolution~\cite{zhu2019deformablev2} in the backbone networks. 
  We achieve the best things-stuff trade-off to get best final PQ results.}
\end{table*}

\begin{table}[t]
  \begin{centering}
  \tabcolsep 0.02in\renewcommand{\arraystretch}{1.1}{\footnotesize{}}%
  \begin{tabular}{c|ccc|c|c}
  \hline 
  Method & PQ & PQ$^{th}$ & PQ$^{st}$ & Type & Cost \tabularnewline
  \hline 
  \hline
  Baseline & 41.0 & 47.2 & 31.5 & - & 1x\tabularnewline
  \hline
  GradNorm & 41.8 & 48.0 & 32.4 & G & $\sim$2x\tabularnewline
  Grid Search & 42.1 & 49.0 & 31.6 & S & $\sim$20x\tabularnewline
  PBT & 41.4 & 49.1 & 29.8 & S & $\sim$8x\tabularnewline
  AM-LFS & 41.7 & 50.1 & 28.9 & R & $\sim$8x\tabularnewline
  BOHB & 42.0 & 50.0 & 29.9 & B & $\sim$8x\tabularnewline
  \hline
  Ada-Segment & \textbf{43.7}$_{\pm0.1}$ & \textbf{51.2}$_{\pm0.07}$ & 
  32.5$_{\pm0.14}$ & R & $\sim$8x+1x\tabularnewline
  \hline 
  \end{tabular}{\footnotesize\par}
  \par\end{centering}

  \caption{\label{tab:Searching-method}
  Comparison with different types of automated tuning methods on COCO \textit{val} split,
  including GradNorm~\cite{chen2017gradnorm}, Grid Search, PBT~\cite{jaderberg2017pbt}, 
  AM-LFS~\cite{li2019AMlfs} and BOHB~\cite{falkner2018bohb},
  on COCO \textit{val} set based
  on our baseline network (R-50 backbone). 
  G: Gradient-Based, S: Sample-Based B: BO-Based, R: RL-guided. 
  We ran Ada-Segment 3 times with different random seeds
  and report in format of mean$\pm$std.}
\end{table}

 \subsection{Ablation Studies}

 
 \noindent
 \textbf{Main Results}.
 We compare with different baselines in Table~\ref{tab:Ablation-Studies-on-coco}. 
 Compared with the vanilla baseline using
 all weights equal to 1, our method achieves 43.7\% PQ, bringing 2.7\% performance gain. 
 One may argue that whether the vanilla baseline is appropriate since we introduce 
 more computational cost and the vanilla weight setting may have bias on different network structures. 
 Therefore, we provide an alternate baseline named \textit{baseline-G} 
 following the way used in~\cite{panopticFPNkirillov2019}, which treats detection losses
 as a single group and performs grid search on detection and segmentation loss weights. 
 It can be seen that our Ada-Segment also achieves 1.5\% performance gains.

 \noindent
 \textbf{Effectiveness of Weight Controller}.
 To validate the effectiveness of the weight controller, in the top part of 
 Table~\ref{tab:Ablation-Studies-on-coco},
 we remove the weight controller to see whether it could work well when
 only synchronize all models at each time checkpoints with different loss weights 
 generated by an evolution strategy. 
 This setting resembles Population Based Training~\cite{jaderberg2017pbt}, 
 and is also used as a baseline named by \textit{baseline-P}. 
 The results suggest that the controller contributes most 
 in the loss weights adaption.

 Besides, we can obtain the best-performed model $M^T_{best}$ after Automated Exploration 
 and Adaptation phase. We also 
 report the performance of $M^T_{best}$ as \textit{Ada-Segment-A} in table~\ref{tab:Ablation-Studies-on-coco}, 
 which can already brings 1.9\% PQ gain compared with the vanilla baseline and outperforms 
 \textit{baseline-G} by 0.9\%. When train model with 
 the controller using policy ensemble, additional 0.6\% performance gain is obtained, demonstrating 
 that the controller did learn the potential training pattern and benefit the model training.

    
 \noindent
 \textbf{Superiority of Training Dynamics}.
 Different from the a static weighting strategy, we leverage the weight controller 
 to dynamically adjust training loss weights. 
 One may have the question that whether it is enough to use the policy network to give a 
 static weight setting that benefits the whole training process. We validate this concern
 by trying the following settings:
 1) \textit{Final}: training with the final loss weights $\boldsymbol{\lambda}_{best}^T$ of the Exploration and Adaptation phase. 
 2) \textit{Pred}: using the final policy network to predict a static loss weights (based on the initial loss) 
 to train the model.  
 3) \textit{Single-dy}: using the final policy network to guide the training process.
 4) \textit{Comb-dy}: using the weight controller with the policy ensemble strategy 
 during the training process. 
 
 The results in Table~\ref{tab:ana-dy} demonstrate the significant advantage of
 dynamically adapting the loss weights during training. 
 Using policy ensemble to combine all states of policy networks improves the 
 final state of policy network $\pi^T$ by 0.6\%. This is not surprising since the policy network 
 is continuous updating in the exploration phase, and the final state may not well suite for 
 the begining stages. It also suggests that there do exists ever-chaning convergence dynamics 
 during panoptic model training.


    
    

 \noindent
 \textbf{Generalization Capability across Network Structures}. 
 The main results of our method are carried our on a well-developed network to show the problem
 that loss weighting inhibits the network-level design. 
 One may curios about whether our framework can also work well on a simple network with less losses.
 To validate the generalization capability of our framework, 
 we apply our Ada-Segment strategy to the simple Panoptic FPN~\cite{panopticFPNkirillov2019},
 which uses Mask R-CNN~\cite{he2017mask} as base detector with three detection losses 
 (bounding box regression / classification, mask segmentation losses) and uses the \textit{same}
 segmentation head as used in our network, 
 and the results are shown in Table~\ref{tab:Evaluation-on-panopticfpn}.
 
It can be seen that although the network contains fewer losses than the network used 
 in our method (4 vs 10 losses), the imbalance problem is also serious since it degrades 0.9\% PQ 
 and 2.3\% PQ$^{th}$. 
 With our Ada-Segment, we boost the performance of both PQ$^{th}$ and PQ$^{st}$ 
 and improve the final performance by 1.8\% PQ from the vanilla weighting and 0.9\% from the well-tuned baseline. 


 \noindent
 \textbf{Policy Transferability}.  
 In Table~\ref{tab:transfer}, we evaluate the transferability of the policy network
 on different backbones, training schedules and datasets.
 It can be found that, 
 1) the policy benefits model training when transfer from ADE20K to COCO.
 2) Policy from COCO also has a positive effect on ADE20K (32.0\% vs. 31.6\% baseline) and 
 comparable with grid-search baseline (32.1\%). 
 3) Policy trained along with small backbone network also works well on larger backbone.
 4) Policy obtained from short training schedule can also be apply on longger training schedule, suggesting
 that the training dynamics of different training schedules are similar. 

 \noindent
 \textbf{Reward Function Design}.
 When only using $r_{local}$ as rewards, the controller learns from the relative differences between 
 sampled actions but overlooks the training dynamics between checkpoints, which only get 
 42.6\% PQ finnally.
 With $r_{imp}$, the controller is trained much well to get 43.7\% PQ. 

\subsection{Comparison with other methods}

\noindent
\textbf{Panoptic Segmentation on COCO}.
We compare our proposed network with other state-of-the-art methods
on COCO \textit{test-dev} split in Table~\ref{tab:Results on COCO test-dev}. 
With the proposed method, we achieve the PQ performance 48.5\%, which
is the state-of-the-art results produced by a single model without
extra training data. 
It is worth noting that although our method does
not report top performance on neither PQ$^{th}$ nor PQ$^{st}$ on
\textit{test-dev} set, 
we achieve the best things-stuff trade-off to get best final PQ results
with our Ada-Segment to reconcile multiple subtask losses during training while previous works may favor
one of the metrics and degrade another.

\noindent
\textbf{Automated Tuning Methods}.
Our method can be seen as an online hyperparameter tuning framework. 
In Table~\ref{tab:Searching-method}, we compare our Ada-Segment with different types
of automated tuning methods to show the practicability and effectiveness of our method.
Grid search and PBT\cite{jaderberg2017pbt} are used as special baselines of as showed in the previous sections. 
GradNorm~\cite{chen2017gradnorm} is an effective multi-objective learning method 
that adjusts subtask gradients during back-propagation. 
We implement AM-LFS\cite{li2019AMlfs} to directly optimize the loss weights, which 
performs poorly on $PQ^{st}$, which suggests that our weight controller perfors much 
better than the greedy optimization.

Besides, when treating the loss weights as hyperparameters, we compare 
with BOHB~\cite{falkner2018bohb}, the state-of-the-art hyperparameter
optimization method, which performs much worse than our automated adaptation strategy since 
it only searches for a static parameter setting, missing the chance to adjust losses 
at different training stages. 

\begin{figure}[t]
  \centering    
  \includegraphics[width=1\linewidth]{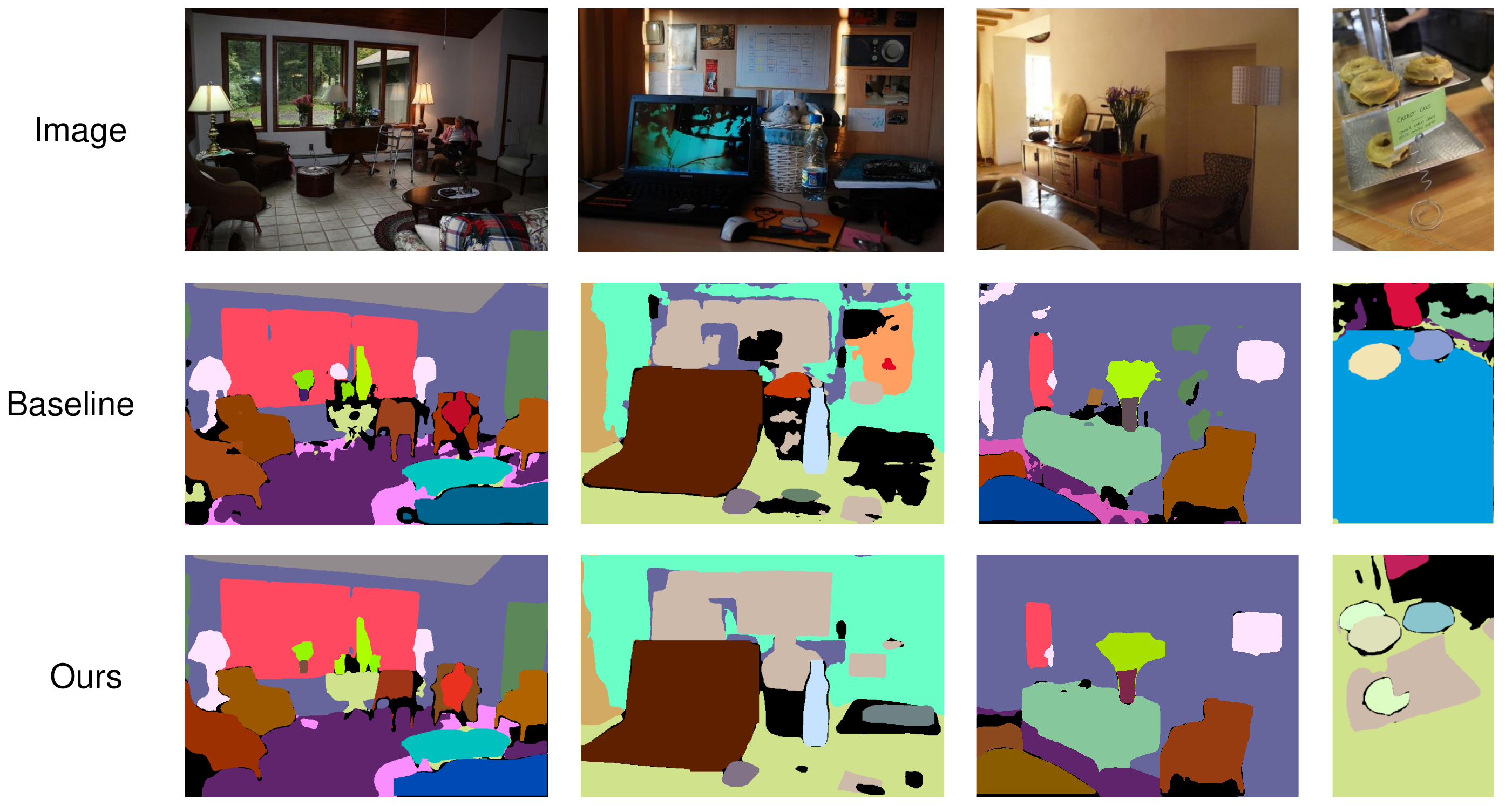}    
  \caption{Qualitative comparison of the results produced by our baseline and 
  the results using Ada-Segment for training. }
  \label{visualization}   
\end{figure}

 \subsection{Qualitative Results}
 In Figure~\ref{visualization}, the results produced by our method are visually precise and coherent
 for both foreground objects and background stuff and the results output by our baseline contain 
 some fuzzy part due to inappropriate training. 
 More qualitative results can be found in the Appendix.

 
 \section{Conclusion}
 In this work, we propose a novel automated online multi-loss adaptation framework named Ada-Segment 
 for panoptic segmentation. We emphasize the importance of dynamically adjusting 
 the loss weights and propose the online multi-loss adaptation strategy with an effective 
 and efficient weight controller, which achieves state-of-the-art performances
 on COCO and ADE20K panoptic segmentation benchmarks. 
 We hope our work will give researchers in this area new insights to focus on the training level design. 

\textbf{Acknowledgments.} 
This work was supported in part by National Natural Science Foundation of China (NSFC) under Grant No.U19A2073 and No.61976233, Guangdong Province Basic and Applied Basic Research (Regional Joint Fund-Key) Grant No.2019B1515120039, Nature Science Foundation of Shenzhen Under Grant No. 2019191361, Zhijiang Lab’s Open Fund (No. 2020AA3AB14) and CSIG Young Fellow Support Fund.  

\bibliography{aaai21}

\end{document}